\pdftrailerid{}
\documentclass[11pt]{article}
\usepackage[preprint]{acl}

\usepackage{times}
\usepackage{latexsym}
\usepackage[T1]{fontenc}
\usepackage[utf8]{inputenc}
\usepackage{microtype}
\usepackage{inconsolata}
\usepackage{graphicx}
\usepackage{booktabs}
\usepackage{placeins}
\usepackage{stfloats}
\usepackage{array}
\usepackage{amsmath}

\newcommand{\dci}[2]{\begin{tabular}[t]{@{}c@{}}#1\\#2\end{tabular}}
\newcommand{\tci}[3]{\begin{tabular}[t]{@{}c@{}}#1\\#2\\#3\end{tabular}}
\let\origtextsc\textsc
\renewcommand{\textsc}[1]{\textup{\origtextsc{#1}}}
\renewcommand{\dbltopfraction}{0.85}
\title{Asking What a Director Sees Can Change\\
What Vision-Language Models Choose}

\author{Jiaxuan Guo$^{1}$\quad Xingyu Ren$^{2}$\quad Kejia Zhang$^{3}$\quad
Jingxin Yang$^{1}$\quad Youran Sun$^{3}$\quad Haizhao Yang$^{3}$\\
$^{1}$Stanford University\quad $^{2}$The Chinese University of Hong Kong\\
$^{3}$University of Maryland, College Park\\
\texttt{guojx@stanford.edu}}

\begin{document}
\maketitle

\begin{abstract}
Vision-language models are now routinely tested on theory of mind: whether they can work out what another person sees, knows, and means.
A common design asks the model to report that perspective and then to act for the person, and reads a correct report beside a wrong action as a failure to use the perspective.
That reading rests on a quiet assumption: that asking for the report leaves the action unchanged.
We show that it does not.
In a rendered Director Task, a partition hides the largest block from a director who asks for ``the largest block''.
Asking first which blocks the director can see raises the share of replies that choose this hidden block by 12 to 88 percentage points on six GPT models and by 17 on Qwen3.8-27B, even though the models mostly report correctly which blocks the director sees.
Reasoning traces and prompt rewordings suggest why: the question shifts how the request is read, from the director's view toward the model's own.
When the prompt states what the director knows and refers to, fewer than 1\% of replies pair a correct report with a hidden-block choice.
Any evaluation that elicits a perspective report before the action it scores needs a baseline without the question, and we deliver one: paired prompts that differ only in that question, a design any perspective-taking benchmark can adopt.
\end{abstract}

\section{Introduction}
\label{sec:intro}

Vision-language models are increasingly asked to work alongside people: to fetch the object someone names when the name only fits from that person's position, or to answer a question about a scene that looks different from the other side of the table.
None of this can be done from the model's own view.
It rests on one ability: working out what another person sees, knows, and means.
Perspective-taking is what turns a model that can look at a scene into a model that can help someone in it, and theory-of-mind tests for vision-language models put it at the centre.
Because the ability is this important, its reported failures matter.
Many tests ask for the report and the action in one prompt: the model states what the other person can see or know, then predicts what that person will do, or acts for them \citep{gu2024simpletom,juneja2026enacttom,zhang2026crosstask}.
When the report is right and the action is wrong, the pair is read as a failure to use the perspective: the model can see the other's view but does not work from it.
If that reading holds, models fail at exactly the skill that working with people requires.
The verdict has become one of the standard findings about vision-language models.

The Director Task tests this use of a perspective in people.
A director and a participant sit on two sides of a grid of slots, some blocked from the director's view, and the procedure makes clear that only objects both of them can see are part of the game \citep{keysar2000taking}.
When the director asks for ``the small candle'' and the smallest candle stands in a blocked slot, reaching for that candle is scored as an egocentric error, and adults still make it \citep{keysar2000taking,keysar2003limits}.

The verdict rests on an assumption that the standard protocol never checks: asking for the report leaves the action unchanged.
There are reasons to doubt it.
In the closest prior Director Task design for vision-language models, asking first which blocks the director can see raised the mean share of correct moves over nine models from 12.0\% to 59.3\% \citep{zhang2026crosstask}.
In people, asking whether someone will do something changes what they do \citep{sherman1980self,morwitz1993measuring,spangenberg2016meta}, and survey respondents read a question in the light of the questions around it \citep{schwarz1999self}.

The difficulty is that the effect of asking hides inside the standard protocol.
A test that elicits the report and the action together never shows what the action would have been without the report question, and the effect has no fixed direction: in that prior design asking made correct moves more common, and in ours it makes choices of the hidden block more common (\S\ref{sec:asking}).
Without a baseline, a correct report beside a wrong action cannot tell a failure of the model from an effect of the test.

\begin{figure*}[t]
\centering
\includegraphics[width=\textwidth]{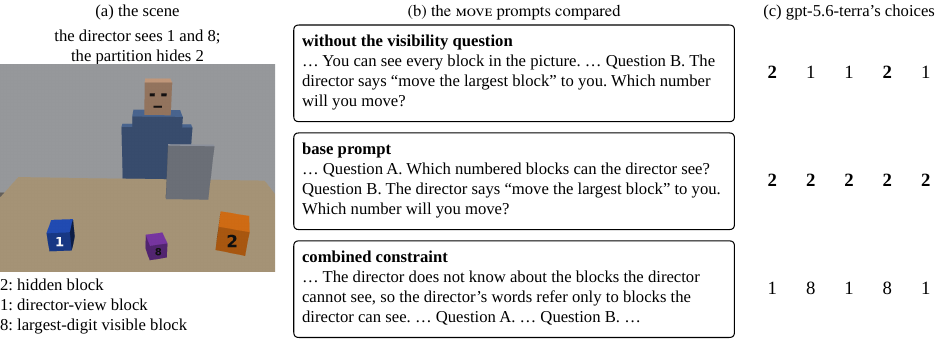}
\caption{One scene, the first in scene order in which the largest block the director can see also looks the largest of them from where the director stands.
(A nearer block can look larger than a bigger one farther away; Appendix~\ref{app:stim}.)
(a)~The partition hides block~2, the largest block from the model's viewpoint.
Of the blocks the director sees, 1 and 8, block~1 is the largest: the director-view block and the keyed answer.
(b)~The three \textsc{move} prompts compared, abbreviated from Table~\ref{tab:prompt}.
(c)~gpt-5.6-terra's block choice in each of its five replies to each prompt, the hidden block in bold.
Without the question it chooses the hidden block in two of five replies on this scene, more often than over all scenes (Figure~\ref{fig:asking}).
The first two rows come from one batch and the third from another (\S\ref{sec:design}).}
\label{fig:example}
\end{figure*}

We measure the effect in a rendered Director Task (Figure~\ref{fig:example}).
A director faces the model across a table of numbered blocks, and an opaque partition hides from the director the block that is largest from the model's viewpoint, the \emph{hidden block}.
The \emph{visibility question} asks which blocks the director can see.
A second question then asks either which block the director means by ``the largest block'' (\textsc{mean}), or which block the model will move when the director says ``move the largest block'' (\textsc{move}).
Unlike the human task, the prompt states only the rule for when the director can see a block, not what the director knows or which blocks the director's words refer to.
Each comparison pairs two prompts that differ in one part and otherwise match byte for byte, so whatever changes in the answers belongs to that part.

Our hypothesis is that asking the visibility question does more than collect a report: it changes how the model reads the request that follows.
In a request about its own move, the model then reads ``the largest block'' as the largest block in its own view, which is the hidden block here.
Figure~\ref{fig:logic} lays out how each comparison tests this hypothesis or an alternative account that would also raise choices of the hidden block (\S\ref{sec:design}).
The results tell one story, from symptom to explanation to fix.

\begin{itemize}
\item \textbf{Symptom: asking for the report moves the action.}
Adding the visibility question raises the share of \textsc{move} replies that choose the hidden block by 12 to 88 points on six GPT models and by 17 on Qwen3.8-27B.
Under \textsc{mean} the rise is far smaller, and the answers to the visibility question are about as accurate under \textsc{mean} as under \textsc{move} (\S\ref{sec:asking}).
The action a test scores can thus depend on the test's own questions.
\item \textbf{Explanation: the question changes how the request is read.}
With the question, the reasoning of Qwen3.8-27B under \textsc{move} takes the largest block in its own view in 39\% of its traces instead of 12\%.
These traces state the director's view but set it aside.
Rewordings of the request move most of the GPT models as the hypothesis predicts: fewer hidden-block choices when the request names what the director wants, more when it names the model's own move (\S\ref{sec:readings}).
A model can then be answering a question about its own view rather than the one the test meant to ask.
\item \textbf{Fix: state what the director knows and refers to.}
Once the prompt spells this out, fewer than 1\% of replies pair a correct report with a hidden-block choice, and asking hardly raises choices of the hidden block.
Most choices that still miss the largest block the director can see take the visible block with the largest digit (\S\ref{sec:constraint}).
What the base prompt left open was how to read the request.
\end{itemize}

A correct report beside a wrong action therefore does not by itself license the verdict that the model failed to use the perspective.
It raises a question that the standard protocol cannot answer: was that the model, or the question?
We answer it for this task, and give the measurement that answers it for others: prompts paired with and without the question.

\section{Related Work}
\label{sec:related}

\paragraph{What the field measures.}
Theory-of-mind benchmarks for language models mostly score predictions of another agent's behaviour, and good predictions need not lead to good actions \citep{riemer2025position,zhou2023far}.
SimpleToM \citep{gu2024simpletom} asks what a character is aware of apart from what the character will do.
GPT-4o answers the first far more accurately than the second, and it answers the second more accurately when the prompt supplies the first question together with its answer.
Related benchmarks ask for belief and action separately \citep{gandhi2023bigtom}, check whether answers in several formats agree \citep{kim2023fantom,li2024benchmarking}, or score who perceives each piece of information apart from the beliefs inferred from it \citep{jung2024perceptions}.
Agent benchmarks elicit a belief or a mapping beside the action \citep{juneja2026enacttom,goel2026bayesian,wang2026seeing}, or score what each sub-agent needs both in a mapping and in the prompts written for it \citep{sun2026perspectivegap}.
Two prompting moves help a model along: keep only what a character knows before answering \citep{wilf2024think}, or make explicit what the character does not know \citep{jeon2026picture}.
Across these designs the report and the action share a prompt or an output, and the pair is read as evidence about the model.
The present task follows that pattern: the model writes the report itself, in the reply that gives its own move.

\paragraph{The Director Task in people.}
Adults make more errors following a director than following a rule that excludes the same objects \citep{dumontheil2010online,apperly2010limits}, and a speaker can refer to objects outside the speaker's current view \citep{rubiofernandez2017director}.
What a speaker can see thus does not settle what the speaker's words refer to, and the constraint of \S\ref{sec:constraint} states both.

\paragraph{Wording moves scores.}
Benchmarks test level-1 visual perspective-taking, judging what another person can see, and level-2, judging how it looks to them \citep{flavell1981young}, in vision-language models \citep{gao2024vision,goral2025beyond,wang2026egocentric,linsley2025threedpc}.
In a Director Task, GPT-4o's accuracy changes with whether the instruction says ``from your point of view'' or ``from my point of view'' \citep{prunty2026visuospatial}.
When a prompt states no frame of reference, vision-language models mostly take the camera's \citep{zhang2025comfort}.
Small changes to a task's content, wording or format move theory-of-mind and other scores \citep{ullman2023trivial,sclar2023minding,shapira2024clever,sclar2024quantifying,mizrahi2024state}, and they also move models' answers to survey questions \citep{tjuatja2024llms,dominguezolmedo2024questioning}.
A constrained-option score can end up reading a model that is answering a different question \citep{guo2026scoring}.
The field therefore knows that prompts move scores.
What these studies leave aside is the one prompt element every report-before-action test shares: the act of asking for the report.
The closest prior design varies it, and there asking reduced errors (\S\ref{sec:intro}).

\section{Task and Design}
\label{sec:design}

\paragraph{Scenes.}
Each scene shows a tabletop from the model's viewpoint, with three or four numbered cubes, an opaque partition and a director figure facing the model.
The partition hides exactly one block from the director.
In the 160 scenes of the \emph{main set}, the hidden block is the largest block, the cube with the longest edge.
A choice of the hidden block is thus a choice of the largest block in the model's own view.
Each cube's edge is at least 1.20 times that of the next smaller cube.
Digits are drawn at random, independently of size.
An independent renderer recomputes every answer key (Appendix~\ref{app:stim}).
Results are on the main set unless a table or figure names another set.
The \emph{rotated set} serves one check in \S\ref{sec:asking}: in its scenes the partition hides the largest, the middle or the smallest of three blocks.

\begin{table*}[t]
\centering
\footnotesize
\setlength{\tabcolsep}{5pt}
\renewcommand{\arraystretch}{0.95}
\begin{tabular}{@{}>{\raggedleft\arraybackslash\itshape}p{0.13\textwidth}|>{\ttfamily}l@{}}
\toprule
 & The picture shows a tabletop from where you are sitting. A director figure \\
 & stands on the far side of the table, facing you. An opaque vertical partition \\
 & stands on the table. Every block is a cube with one digit painted on all four \\
 & of its vertical faces; that digit is the block\textquotesingle s number. \\[3pt]
 & The director can see a block only when nothing stands between that block and \\
 & the director\textquotesingle s eyes. The partition is opaque and nothing can be seen through \\
 & it. You can see every block in the picture. \\[3pt]
inserted paragraph & \textrm{\itshape (none in the base prompt)} \\[3pt]
 & Answer with block numbers. A colour word is never a valid answer. \\[3pt]
Question A & Question A. Which numbered blocks can the director see? \\[3pt]
Question B & Question B. The director says "move the largest block" to you. Which number \\
 & will you move? \\[3pt]
 & Reply with one JSON object and nothing else, in exactly this shape: \\
reply format & \{"visible\_to\_director": [<block numbers>], "chosen\_block": <one block number>\} \\
\bottomrule
\end{tabular}

\vspace{6pt}
\scriptsize
\setlength{\tabcolsep}{4pt}
\begin{tabular}{@{}>{\raggedright\arraybackslash}p{0.14\textwidth}>{\raggedright\arraybackslash}p{0.82\textwidth}@{}}
\toprule
part & variants, under the names that the text and the tables use \\
\midrule
Question~A & \emph{visibility question}, as above; \emph{without the visibility question}: the question, the blank
line after it and its reply field removed; \emph{block-count question}: ``How many blocks are on the table?'',
answered in the field \texttt{block\_count}; \emph{knowledge question}: ``Which numbered blocks does the
director know are on the table?'', answered in the field \texttt{visible\_to\_director}; \emph{visible blocks
stated}: the line ``The director can see blocks \dots{} and no other block.'' in place of the question, with
the choice as the only reply field; \emph{question after the request}: the question after Question~B, its field
after \texttt{chosen\_block}; \emph{question in a first turn}: the question and its field alone in a first
turn, and Question~B with \texttt{chosen\_block} in a second turn of the same conversation \\
Question~B & \textsc{move}, as above; \textsc{mean}: ``The director says "the largest block" to you. Which number
does the director mean?''; \textsc{move} with its second line reworded: \emph{``want you to move''}: ``Which
number does the director want you to move?'', or \emph{``the block that you will move''}: ``Which number is on
the block that you will move?'' \\
inserted paragraph & \emph{combined constraint}: ``The director does not know about the blocks the director
cannot see, so the director's words refer only to blocks the director can see.''; \emph{knowledge
constraint} and \emph{reference constraint}: its first and its second clause alone, each padded to the same
length; \emph{padding sentence}: ``The picture was drawn by a computer program written for this task.'' \\
\bottomrule
\end{tabular}
\caption{Top: the base prompt with \textsc{move}, sent with the scene image, line breaks as sent; the left column marks the parts that conditions change.
Bottom: the variants of each part (Appendix~\ref{app:prompt}).}
\label{tab:prompt}
\end{table*}

\paragraph{Prompts.}
The prompt describes the scene and asks two questions, labelled Question~A and Question~B (Table~\ref{tab:prompt}).
The reply gives both answers in one JSON object.
In the \emph{base prompt}, Question~A is the visibility question and Question~B is \textsc{mean} or \textsc{move}.
Each comparison changes one part of the prompt to one of the variants in Table~\ref{tab:prompt} and leaves the rest identical byte for byte (Appendix~\ref{app:prompt}).
A \emph{condition} is one prompt asked of one model.

\paragraph{Batches.}
Conditions that we compare with each other were sampled together, in one random order per model.
We call such a set of conditions a \emph{batch} (Table~\ref{tab:design} in Appendix~\ref{app:prompt}).
Each batch includes the base prompt.
Across the batches that ask \textsc{move}, the share of a model's replies to the base prompt that choose the hidden block varies by up to 8.1 points (gpt-6-sol; Table~\ref{tab:batches}).
Every comparison therefore sets conditions of one batch against each other.

\paragraph{Measures.}
A reply's answer to the visibility question is its \emph{visibility report}.
The report is correct when it lists exactly the blocks the director can see.
\emph{Visibility accuracy} is the share of answered replies with a correct visibility report.
The keyed answer to Question~B is the \emph{director-view block}, the largest block the director can see.
A \emph{hidden-block choice} names the hidden block, whatever the report says.
The \emph{hidden-block choice rate} is the share of answered replies that make one.
Because this rate does not need a report, it can compare conditions with and without the visibility question.
A hidden-block choice beside a correct visibility report is the pair that tests of perspective-taking read as a failure to use the report.
The \emph{joint rate} is the share of answered replies that hold such a pair.
The \emph{largest-digit visible block} is the block with the largest digit among those the director can see, counted only in scenes where it is not the director-view block.

\paragraph{Models.}
We test seven GPT models \citep{openai2026gpt55card,openai2026gpt56card,openai2026gpt6astracard}, grok-4.6 \citep{xai2026grok46card} and two open-weight models served locally, Qwen3.8-27B \citep{qwen2026qwen38card} and Muse Glimmer-30B \citep{meta2026museglimmercard}.
All answer at medium reasoning effort (Appendix~\ref{app:prompt}).
Within GPT-5.6, Sol is the provider's flagship, Terra a lower-cost option and Luna its fastest and most cost-efficient model \citep{openai2026gpt56card}.
Within GPT-6, Astra is the provider's most capable broadly deployed model, Sol a lower-cost alternative to it and Luna its fastest and most cost-efficient model \citep{openai2026gpt6astracard}.
gpt-6-astra was asked only with the visibility question.
We call the other GPT models \emph{the six GPT models}.
Table~\ref{tab:design} shows which models were asked each comparison.

\paragraph{Analysis.}
Each condition asks each scene five times, for 800 calls per condition.
Differences between conditions are reported in percentage points (\emph{points}), with 95\% intervals from generalized estimating equations with scenes as clusters \citep{liang1986longitudinal}.
We call a difference an increase or a decrease only when its interval excludes zero and it survives Benjamini--Hochberg correction within its section \citep{benjamini1995controlling}.
Tables mark such differences with an asterisk (Appendix~\ref{app:stats}).

\begin{figure*}[t]
\centering
\includegraphics[width=\textwidth]{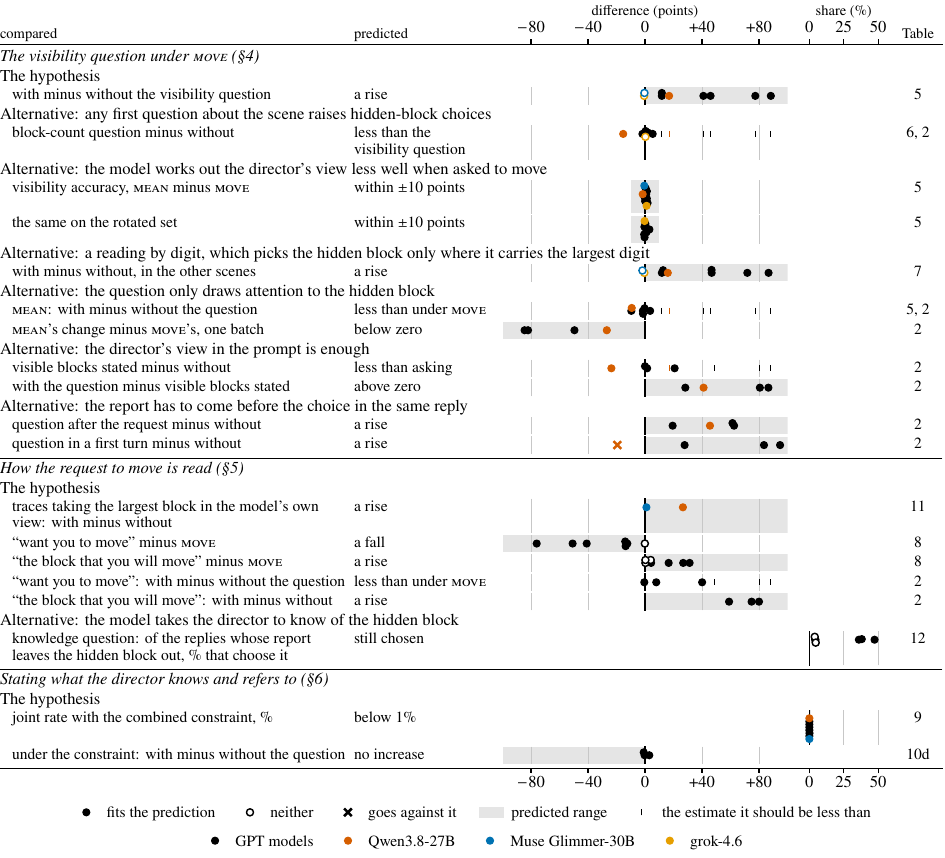}
\caption{The comparisons of \S\S\ref{sec:asking}--\ref{sec:constraint}, one per line, and what the hypothesis predicts.
Each mark is one model's estimate, in points or, in the right-hand panel, in percent: filled where it fits the prediction, hollow where it does neither and a cross where it goes against it, by the rule of \S\ref{sec:design} or, for the knowledge question, as \S\ref{sec:readings} explains.
Grey shading marks the predicted range, and a short vertical line, in a model's colour, the estimate that the model's mark should be less than.
The GPT models are black, Qwen3.8-27B vermillion, Muse Glimmer-30B blue and grok-4.6 orange.
The right column names the tables that hold the estimates; variants: Table~\ref{tab:prompt}.}
\label{fig:logic}
\end{figure*}

\section{The Visibility Question Raises Hidden-Block Choices under \textsc{move}}
\label{sec:asking}

\begin{figure}[t]
\centering
\includegraphics[width=\columnwidth]{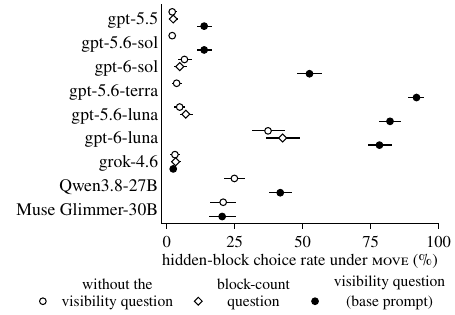}
\caption{Shares of the answered \textsc{move} replies that choose the hidden block, with 95\% intervals over scene resamples: without the visibility question, with the block-count question in its place where this batch asks it (gpt-5.6-terra and Qwen3.8-27B were asked it in the batch of Table~\ref{tab:e13}), and with the visibility question, the base prompt.
gpt-6-astra, asked only with the visibility question, chooses the hidden block in 2 of 800 replies and is not shown.
Counts: Table~\ref{tab:askcounts}.}
\label{fig:asking}
\end{figure}

\begin{figure*}[t]
\centering
\includegraphics[width=\textwidth]{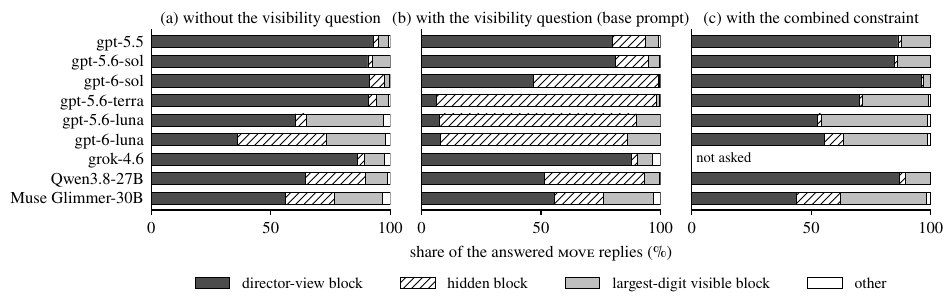}
\caption{Block choices as shares of the answered \textsc{move} replies: (a)~without and (b)~with the visibility question, from one batch per model, and (c)~with the combined constraint, for GPT models from another.
Numbers: Table~\ref{tab:choices}.}
\label{fig:choices}
\end{figure*}

\paragraph{The visibility question raises hidden-block choices under \textsc{move}.}
Asking the question raises the hidden-block choice rate under \textsc{move} by 12 to 88 points on the six GPT models, and by 17 points on Qwen3.8-27B (Figure~\ref{fig:asking}, Table~\ref{tab:summary}).
Without the question, each of these GPT models except gpt-6-luna chooses the hidden block in at most 7\% of \textsc{move} replies.
The action that a test scores thus changes when the test asks for the report.
With the question, fewer replies choose the director-view block on each of these models and on Qwen3.8-27B (Figure~\ref{fig:choices}).
On grok-4.6 and Muse Glimmer-30B the interval for the change includes zero.

\paragraph{A question about the number of blocks raises the rate far less than the visibility question.}
On the five GPT models asked both, the block-count question of Table~\ref{tab:prompt} took the place of the visibility question.
This control tests whether any first question has the effect, since answering several questions from different sources in one reply can already change the answers \citep{wang2024evaluating}.
On these models the block-count question changes the hidden-block choice rate under \textsc{move} by at most 5.4 points (Tables~\ref{tab:e13} and~\ref{tab:askcounts}), while in the same batches the visibility question raises the rate by 12 to 88 points.
Answering a first question about the scene thus accounts for a small part of the increase at most.
On grok-4.6 neither question changes the rate.
On Qwen3.8-27B the block-count question lowers it, by 15.3 points.

\paragraph{Visibility accuracy differs little between \textsc{mean} and \textsc{move}.}
If the model worked out the director's view less well when asked to move, its visibility reports would be less accurate under \textsc{move} than under \textsc{mean}.
They are not.
We take visibility accuracy as unchanged when the interval of this difference lies within $\pm$10 points, a change of one reply in ten (Appendix~\ref{app:stats}).
Under the base prompt the interval lies within $\pm$4.4 points on the main set and within $\pm$5.8 points on the rotated set (Table~\ref{tab:summary}).
The second question thus changes the block choice and leaves visibility accuracy within the margin.

\paragraph{The increase appears whether or not the hidden block carries the largest digit.}
A reading of ``the largest block'' by digit would also pick the hidden block, but only where the hidden block carries the largest digit.
In the scenes where it does not, the visibility question still raises hidden-block choices by 12 to 87 points on the six GPT models and by 16 on Qwen3.8-27B (Table~\ref{tab:strata}).
It raises them in the other scenes as well (Table~\ref{tab:strata}).

\paragraph{Under \textsc{mean}, the visibility question raises hidden-block choices by at most 3.6 points.}
If the question only drew attention to the hidden block, it would raise hidden-block choices under \textsc{mean} about as much as under \textsc{move}; an instruction that singles out an object does make speakers mention it more often \citep{lane2006dont}.
The question does not.
In the batch of Table~\ref{tab:e13}, gpt-5.6-terra, gpt-6-sol and gpt-5.6-luna were asked both questions with and without the visibility question.
With it, they choose the hidden block at most 3.6 points more often under \textsc{mean}, and 49 to 88 points more often under \textsc{move}.
In the same batch the question lowers Qwen3.8-27B's hidden-block choices under \textsc{mean}.
In a batch that asks \textsc{mean} alone, it raises them on none of gpt-5.5, gpt-6-sol, gpt-5.6-luna and gpt-6-luna (Table~\ref{tab:summary}).

\paragraph{Asking raises hidden-block choices far more than stating which blocks the director can see.}
The increase might need only the director's view in the prompt, not the question.
To test this, one batch on gpt-5.6-terra, gpt-6-sol, gpt-5.6-luna and Qwen3.8-27B states the visible blocks in place of the visibility question (Table~\ref{tab:prompt}).
On all four models the question raises hidden-block choices by 28 to 87 points more than the statement does (Table~\ref{tab:e13}).
The statement itself raises them only on gpt-6-sol, by 20.8 points.
On Qwen3.8-27B it lowers them.

\paragraph{On gpt-5.6-terra, gpt-6-sol and gpt-5.6-luna, the question also raises hidden-block choices after the request or in a first turn.}
The rise might need the report to come before the choice in the same reply.
On these three models, it does not.
Asked after Question~B, with the report after the choice in the reply, the question raises them by 19 to 63 points (Table~\ref{tab:e13}).
Asked alone in a first turn, with Question~B in a second turn of the same conversation, it raises them by 28 to 95 points.
On Qwen3.8-27B the question raises them after the request and lowers them in a first turn.
In this batch, each interval for visibility accuracy with the question moved, or under \textsc{mean}, minus the base prompt lies within $\pm$5.1 points.

\section{How the Request to Move Is Read}
\label{sec:readings}

\begin{table*}[t]
\centering
\scriptsize
\setlength{\tabcolsep}{3pt}
\begin{tabular}{@{}>{\raggedright\arraybackslash}p{0.28\textwidth}cccc@{}}
\toprule
 & gpt-5.6-terra & gpt-6-sol & gpt-5.6-luna & Qwen3.8-27B \\
\midrule
with minus without the visibility question & $+88.1$ $[+85.3, +91.0]^{*}$ & $+49.0$ $[+43.7, +54.3]^{*}$ & $+80.6$ $[+77.0, +84.3]^{*}$ & $+17.5$ $[+12.7, +22.3]^{*}$ \\
block-count question minus without & $+0.4$ $[-2.0, +2.7]$ & -- & -- & $-15.3$ $[-19.1, -11.4]^{*}$ \\
visible blocks stated minus without & $+1.5$ $[-0.9, +3.9]$ & $+20.8$ $[+15.9, +25.6]^{*}$ & $0.0$ $[-1.7, +1.7]$ & $-23.6$ $[-27.5, -19.8]^{*}$ \\
with the question minus visible blocks stated & $+86.6$ $[+83.6, +89.7]^{*}$ & $+28.3$ $[+23.0, +33.5]^{*}$ & $+80.6$ $[+77.0, +84.2]^{*}$ & $+41.1$ $[+37.0, +45.3]^{*}$ \\
question after the request minus without & $+62.5$ $[+58.1, +66.9]^{*}$ & $+19.4$ $[+14.8, +23.9]^{*}$ & $+61.4$ $[+56.0, +66.8]^{*}$ & $+45.6$ $[+41.0, +50.3]^{*}$ \\
question in a first turn minus without & $+83.5$ $[+80.2, +86.8]^{*}$ & $+27.8$ $[+22.8, +32.7]^{*}$ & $+94.8$ $[+93.1, +96.5]^{*}$ & $-19.4$ $[-23.1, -15.7]^{*}$ \\
\textsc{mean}: with minus without the question & $+3.6$ $[+0.6, +6.7]^{*}$ & $-0.5$ $[-1.4, +0.4]$ & $-1.5$ $[-3.0, -0.03]^{\dagger}$ & $-9.3$ $[-11.8, -6.7]^{*}$ \\
\textsc{mean} change minus \textsc{move} change & $-84.5$ $[-88.7, -80.3]^{*}$ & $-49.5$ $[-54.7, -44.3]^{*}$ & $-82.1$ $[-85.8, -78.5]^{*}$ & $-26.8$ $[-32.1, -21.4]^{*}$ \\
``want you to move'': with minus without the question & $+7.9$ $[+4.9, +10.9]^{*}$ & $-0.5$ $[-1.0, -0.01]^{\dagger}$ & $+40.1$ $[+36.3, +43.9]^{*}$ & -- \\
``the block that you will move'': with minus without the question & $+80.0$ $[+76.5, +83.5]^{*}$ & $+59.0$ $[+54.6, +63.4]^{*}$ & $+74.8$ $[+71.1, +78.4]^{*}$ & -- \\
\bottomrule
\end{tabular}
\caption{Changes in the hidden-block choice rate under \textsc{move}, or under the second question named, in points with 95\% intervals, each model's conditions in one batch.
$^{*}$:~an increase or a decrease by the rule of \S\ref{sec:design}; $^{\dagger}$:~the interval excludes zero but falls short of the correction (Appendix~\ref{app:stats}); interval ends that round to zero are given to two places; --:~not asked.}
\label{tab:e13}
\end{table*}

\begin{figure*}[t]
\centering
\includegraphics[width=\textwidth,trim=0 10 0 0,clip]{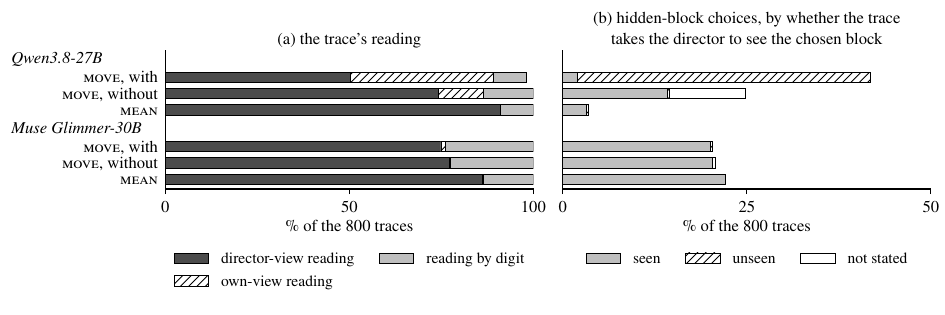}
\caption{The reasoning traces of the open-weight models, as percentages of the 800 traces of each condition: (a)~traces with the director-view reading, the own-view reading or a reading by digit; (b)~hidden-block choices, by whether the trace takes the director to see the chosen block.
Counts and differences: Table~\ref{tab:traces}.}
\label{fig:readings}
\end{figure*}

\paragraph{Reasoning traces.}
Qwen3.8-27B and Muse Glimmer-30B return the reasoning they write before replying, their \emph{reasoning traces}.
An automatic coder, gpt-6-astra at maximum reasoning effort, codes each trace against a codebook (Appendix~\ref{app:traces}).
The coder sees only the codebook, a neutral identifier, the trace and the reply.
Its first code is the trace's reading of ``the largest block'', the rule by which the trace picks its block.
The \emph{director-view reading} takes the largest block among those the trace takes the director to see.
The \emph{own-view reading} takes the largest block among all the blocks as the model sees them.
A reading by digit takes the block with the largest digit.
A second code records whether the trace takes the director to see the chosen block: seen, unseen or not stated.
Where the trace states no belief about that block, this code follows the reply's visibility report.
A third code records whether the trace works out which blocks the director can see.
The coder gives the intended codes on all 30 test traces written to a known code.
Two people unaware of the hypotheses coded 36 sampled traces on their own, and their reading codes agree with the coder's on 36 and 34 of them.

\paragraph{With the visibility question, Qwen3.8-27B's reasoning more often takes the largest block in its own view.}
With the question, 39\% of its \textsc{move} traces take the own-view reading (Figure~\ref{fig:readings}).
Without the question 12\% do, and under \textsc{mean} one trace in 800 does.
The first such trace in scene order is on scene main-002, where the reply chooses the hidden block.
It reads \emph{The phrasing "the largest block" likely means the largest among all blocks, which is 2.}

\paragraph{With the question, Qwen3.8-27B's hidden-block choices come with traces that say the director cannot see the block.}
Every \textsc{move} trace with the question works out which blocks the director can see (Table~\ref{tab:traces}).
In 318 of the 334 hidden-block choices, the trace takes the director not to see the chosen block.
A second coding pass, shown the trace without the reply's visibility report, gives every one of these choices the same code (Appendix~\ref{app:traces}).
Without the question, 114 of the 199 hidden-block choices come with a trace that takes the director to see the block, and 83 with a trace that leaves it open.
With the question, the hidden-block choice is thus made with the director's view stated and set aside.

\begin{figure}[t]
\centering
\includegraphics[width=\columnwidth]{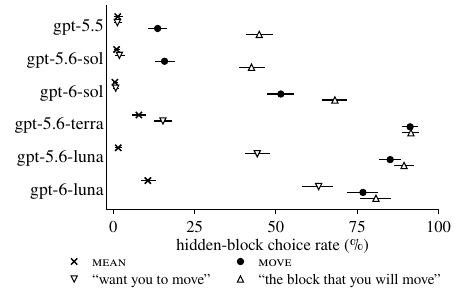}
\caption{Shares of the answered replies that choose the hidden block, with the visibility question, under \textsc{move} (the base prompt), \textsc{mean} and the two rewordings of \textsc{move}, with 95\% intervals over scene resamples.
All four questions come from one batch per model, a different batch from Figure~\ref{fig:asking}'s.
gpt-6-astra chooses the hidden block in at most 4 of 800 replies under each question.
grok-4.6 answered only \textsc{mean} and \textsc{move} in this batch, and the open-weight models were not asked the rewordings.
Counts: Table~\ref{tab:rewordings}.}
\label{fig:rewordings}
\end{figure}

\paragraph{Asking which number the director wants moved lowers the hidden-block choices of the six GPT models.}
These models return no traces, so the evidence comes from two rewordings of the second line of \textsc{move} (Figure~\ref{fig:rewordings}).
Reworded to ``Which number does the director want you to move?'', the line lowers hidden-block choices by 12 to 76 points (Table~\ref{tab:rewordings}).
gpt-5.5, gpt-5.6-sol and gpt-6-sol then choose the hidden block in at most 2\% of replies, as under \textsc{mean}.
gpt-5.6-terra, gpt-5.6-luna and gpt-6-luna still choose it in 15\%, 44\% and 63\% of replies.

\paragraph{Asking which number is on the block the model will move raises them.}
Reworded to ``Which number is on the block that you will move?'', the line raises hidden-block choices on four GPT models, by 4 to 31 points (Table~\ref{tab:rewordings}).
The hidden-block choices of these models thus change with whether the second question names the director's intent or the model's own move.

\paragraph{The visibility question raises hidden-block choices less when the request names what the director wants.}
gpt-5.6-terra, gpt-6-sol and gpt-5.6-luna were asked both rewordings with and without the visibility question, in the batch of Table~\ref{tab:e13}.
Under ``the block that you will move'' the question raises hidden-block choices by 59 to 80 points, and under \textsc{move} in the same batch by 49 to 88.
Under ``want you to move'' it raises them by at most 40.1 points.
This fits the hypothesis, under which asking changes how a request about the model's own move is read.

\paragraph{Asking what the director knows leaves fewer hidden-block choices than asking what the director sees.}
To test whether the hidden-block choices follow a belief that the director knows of the hidden block, the knowledge question of Table~\ref{tab:prompt} replaces the visibility question in one batch per model.
A reply whose own report leaves the hidden block out of what the director knows would not choose that block because of such a belief.
With the knowledge question, each of the six GPT models chooses the hidden block less often (Table~\ref{tab:knowledge}).
Among the replies that leave it out, gpt-6-sol, gpt-5.6-luna and gpt-6-luna still choose it in 36 to 47\% (Table~\ref{tab:knowledge}), so on these three models such a belief does not account for these choices.
gpt-5.5, gpt-5.6-sol and gpt-5.6-terra choose it in at most 5\% of these replies, against 13 to 92\% of the replies whose visibility report leaves it out.
On these models the test does not separate such a belief from the hypothesis.

\section{Stating What the Director Knows and Refers To}
\label{sec:constraint}

\begin{figure}[t]
\centering
\includegraphics[width=\columnwidth]{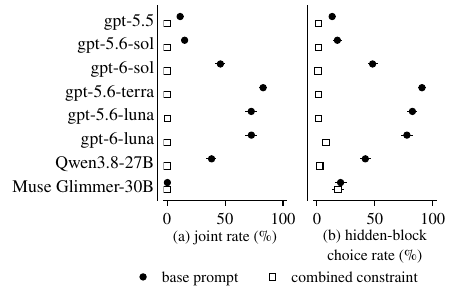}
\caption{\textsc{move} replies under the base prompt and with the combined constraint inserted, in one batch per model: (a)~the joint rate; (b)~the hidden-block choice rate, whatever the report.
Shares of the answered replies with 95\% intervals over scene resamples; most intervals are narrower than the markers.
Counts: Table~\ref{tab:constraintcounts}.}
\label{fig:constraint}
\end{figure}

The \emph{combined constraint} (Table~\ref{tab:prompt}) states what the director knows and what the director's words refer to \citep{brownschmidt2011talking}.

\paragraph{With the combined constraint, the joint rate is below 1\% on every model in every batch.}
Under the base prompt of the same batch, the joint rates of the six GPT models are 11 to 83\% (Figure~\ref{fig:constraint}a).
One batch also asks a copy of the main set in which each scene's digits are rearranged among its blocks, to test whether the joint rate depends on the digits (Appendix~\ref{app:stim}).
The constraint lowers the joint rate on this copy too, and the largest joint rate with it, 0.8\% on gpt-6-luna, comes from it (Table~\ref{tab:parts}c).

\paragraph{The constraint lowers hidden-block choices whatever the report says.}
Under the constraint the six GPT models and Qwen3.8-27B choose the hidden block in at most 7.9\% of replies (Figure~\ref{fig:constraint}b).
The joint rate cannot exceed this share, so its fall is a fall in hidden-block choices.
gpt-6-luna's visibility accuracy also falls under the constraint, by 9 to 15 points in the batches where Appendix~\ref{app:stats} tests this difference.
Muse Glimmer-30B keeps 18.1\% hidden-block choices, all with reports that list the hidden block as visible (Table~\ref{tab:constraintcounts}).

\paragraph{Under the constraint, most choices other than the director-view block take the largest-digit visible block.}
The six GPT models choose the director-view block in 53 to 96\% of \textsc{move} replies under the constraint.
The largest-digit visible block takes 77 to 93\% of their other choices (Table~\ref{tab:choices}).
These choices stay among the blocks the director can see and take the block that a reading of ``the largest block'' by digit picks.

\paragraph{With the constraint, the visibility question hardly raises hidden-block choices.}
gpt-6-sol, gpt-5.6-luna and gpt-6-luna were asked both prompts with and without the question in one batch.
Under the constraint the question changes their hidden-block choice rate by $-0.9$ to $+3.0$ points, and no interval reaches above $+6.2$ (Table~\ref{tab:parts}d).
Under the base prompt it raises the rate by 38 to 81 points.

\paragraph{Each clause of the constraint alone leaves hidden-block choices with correct reports.}
Each clause was inserted alone, padded to the constraint's length.
The first, the knowledge constraint of Table~\ref{tab:prompt}, leaves a joint rate of up to 16.3\% (gpt-5.6-luna; Table~\ref{tab:parts}a).
The second, the reference constraint, leaves up to 10.9\% (gpt-6-luna).
The padding sentence, inserted alone where the constraint stands, lowers gpt-6-luna's joint rate by 13.3 points (Table~\ref{tab:parts}b).
gpt-6-luna thus responds to text that does not concern the task.
On gpt-5.6-luna the interval for this change includes zero.

\section{Discussion}
\label{sec:discussion}

\paragraph{Asking can change which question the model answers.}
Survey research shows how asking reshapes an answer.
When respondents are asked two questions introduced together, first about their relationship and then about their life as a whole, they answer the second as if it left the relationship out \citep{schwarz1991assimilation}.
They follow the conversational norm against repeating what has been said \citep{grice1975logic,strack1988priming,schwarz1999self}.
The visibility question plays the part of that first question, the one about the director's view.
\textsc{move} then asks, in the first person, about the model's own move, after the prompt has told the model that it can see every block.
Read as asking about what the first question left out, \textsc{move} is answered from the model's own view.
\textsc{mean} asks about the director again, the topic the first question covered, so this reading should not appear there, and it hardly does.
A statement of the visible blocks poses no first question at all, so the effect should be smaller after it, and \S\ref{sec:asking} finds that too.
Qwen3.8-27B's traces state the own-view reading outright, and the choices of the six GPT models under the two rewordings fit it (\S\ref{sec:readings}).
On this account the model is not blind to the director's view, since its reports are right; it answers a question that the structure of the prompt created.
An egocentric choice can then come from the conversation rather than from a gap in the model.

\paragraph{The prior design moved the other way.}
Asking first which blocks the director sees made the move more often correct in the closest prior design \citep{zhang2026crosstask}, and less often here.
The two designs differ in their labels and answers (letters and colour words there, digits here), their scenes and their prompt wording, and these data do not say which of these differences sets the direction.
Within this design, what asking does depends on the second question (\S\ref{sec:asking}).
In both designs asking moved the scored action.
An effect that can raise or lower the measured behaviour cannot be wished away as noise; it has to be measured, and the direct way to measure it is the pair with and without the question.

\paragraph{What the community gets.}
Three tools come out of this work.
The first is a baseline design: paired prompts that differ only in the report question, usable in any test that scores an action after asking for a report.
The second is the constraint recipe: state what the other person knows and refers to, so that a choice of an object that person cannot see breaks a stated rule instead of counting as a debatable error (\S\ref{sec:constraint}); here the choices that still miss the keyed block mostly take the visible block with the largest digit.
The third is the reading check: where reasoning traces come back, coding the rule behind a choice shows which question the model answered (\S\ref{sec:readings}).
Each tool is small, and together they turn a confounded pair of scores into a measurement.

\paragraph{Where the field goes from here.}
A correct report beside a wrong action that was measured without this baseline is not worthless, but what it shows stays open until the missing comparison is run.
Once it is run, failures that remain cannot be put down to asking, and those that disappear were effects of asking.
Perspective-taking evaluation can then separate what a model knows about another's view from what the prompt made it do.

\section{Conclusion}

A correct report beside a wrong action is not by itself proof that a model cannot use another's perspective.
In a rendered Director Task, asking a vision-language model which blocks the director can see raises how often it picks the block hidden from the director when it is told to move ``the largest block'', by 12 to 88 points on six GPT models and by 17 on Qwen3.8-27B, although its reports of what the director sees are mostly correct.
When Qwen3.8-27B picks the hidden block after the visibility question, its reasoning mostly says the director cannot see the block and takes the largest block in its own view.
When the prompt adds that the director neither knows about nor refers to blocks the director cannot see, fewer than 1\% of replies pair a correct report with a hidden-block choice.
The question asked for a report can thus change how the request that follows is read, so a test that asks for a report before the action it scores needs a baseline that asks for the same action without the report.

\section*{Limitations}

\begin{itemize}
\item \textbf{One paradigm.}
The scenes come from one rendered family with digits as block labels.
The tests use one referring expression and English prompts.
\item \textbf{Models and effort.}
The measurements cover ten models at medium reasoning effort, and the increase appears on six GPT models and Qwen3.8-27B but is not detected on grok-4.6 or Muse Glimmer-30B; the providers may update the closed models.
\item \textbf{Reasoning traces.}
The reading check applies where models return traces; here that is two open-weight models, and a trace need not be the process that produced the reply \citep{turpin2023language,lanham2023measuring,chen2025reasoning}.
\item \textbf{Asking and having.}
The condition that states the visible blocks in place of asking for them covers four models, and its reply carries no report.
\item \textbf{People.}
The design tests models; whether people change their choices when asked the visibility question remains open.
\end{itemize}

\bibliography{refs}
\clearpage

\appendix
\renewcommand{\textfraction}{0.07}\renewcommand{\topfraction}{0.93}\renewcommand{\dbltopfraction}{0.93}
\renewcommand{\floatpagefraction}{0.85}\renewcommand{\dblfloatpagefraction}{0.85}
\makeatletter\setlength{\@fptop}{0pt}\setlength{\@fpsep}{\floatsep}
\setlength{\@dblfptop}{0pt}\setlength{\@dblfpsep}{\dblfloatsep}\makeatother
\setcounter{topnumber}{3}\setcounter{dbltopnumber}{3}\setcounter{totalnumber}{5}

\section{Prompts, Conditions and Models}
\label{app:prompt}

\paragraph{Variants.}
The variants of Table~\ref{tab:prompt} change the base prompt only as follows.
\textsc{move} has 850 characters against 854 for \textsc{mean}, and each rewording of its second line has 871.
Without the visibility question the prompt is 99 characters shorter, and the label ``Question~B'' stays.
An inserted paragraph stands before the line on block numbers, with a blank line on each side.
The combined constraint has 135 characters and the padding sentence 66.
The knowledge constraint, ``The director does not know about the blocks the director cannot see.'', is padded to the combined constraint's length with the padding sentence.
The reference constraint, ``The director's words refer only to blocks the director can see.'', is padded with ``The picture was drawn by a computer program written just for this task.''
With the visible blocks stated, the line lists the scene's visible blocks in ascending order.
It has 55 characters when the director sees two blocks, as many as the visibility question, and 58 with three.
With the question in a first turn, the second turn carries the first turn's message and reply, and the report comes from the first turn and the choice from the second.

\begin{table*}[t]
\centering
\scriptsize
\setlength{\tabcolsep}{2.2pt}
\newcommand{\hd}[2]{\begin{tabular}[b]{@{}c@{}}#1\\#2\end{tabular}}
\newcommand{\hdd}[3]{\begin{tabular}[b]{@{}c@{}}#1\\#2\\#3\end{tabular}}
\begin{tabular}{@{}r>{\raggedright\arraybackslash}p{0.334\textwidth}l*{10}{c}@{}}
\toprule
row & conditions sampled together & & \hd{}{gpt-5.5} & \hd{gpt-5.6-}{sol} & \hd{gpt-6-}{sol} & \hd{gpt-5.6-}{terra} &
\hd{gpt-5.6-}{luna} & \hd{gpt-6-}{luna} & \hd{gpt-6-}{astra} & \hd{}{grok-4.6} & \hd{Qwen3.8-}{27B} &
\hdd{Muse}{Glimmer-}{30B} \\
\midrule
1 & \textsc{mean}\newline \textsc{move}\newline ``want you to move''\newline ``the block that you will move'' & \S\S4, 5 & \textbullet & \textbullet & \textbullet & \textbullet & \textbullet & \textbullet & \textbullet & \textbullet$^{a}$ & -- & -- \\
2 & \textsc{move}: visibility question\newline \textsc{move}: block-count question\newline \textsc{move}: without the visibility question & \S4 & \textbullet & -- & \textbullet & -- & \textbullet & \textbullet & -- & \textbullet & -- & -- \\
3 & \textsc{move}: visibility question\newline \textsc{move}: without the visibility question & \S4 & -- & \textbullet & -- & \textbullet & -- & -- & -- & -- & -- & -- \\
4 & \textsc{mean}: visibility question\newline \textsc{mean}: without the visibility question & \S4 & \textbullet & -- & \textbullet & -- & \textbullet & \textbullet & -- & -- & -- & -- \\
5 & \textsc{move}: visibility question\newline \textsc{move}: knowledge question & \S5 & \textbullet & \textbullet & \textbullet & \textbullet & \textbullet & \textbullet & -- & -- & -- & -- \\
6 & \textsc{move}: base prompt\newline \textsc{move}: combined constraint & \S6 & \textbullet & \textbullet & \textbullet & \textbullet & \textbullet & \textbullet & -- & -- & -- & -- \\
7 & \textsc{move}: base prompt\newline \textsc{move}: knowledge constraint\newline \textsc{move}: reference constraint & \S6 & \textbullet & \textbullet & \textbullet & \textbullet & \textbullet & \textbullet & -- & -- & -- & -- \\
8 & \textsc{move}: base prompt\newline \textsc{move}: padding sentence\newline \textsc{move}: combined constraint & \S6 & -- & -- & -- & -- & \textbullet & \textbullet & -- & -- & -- & -- \\
9 & \textsc{move}: base prompt and combined constraint, each with and without the visibility question & \S6 & -- & -- & \textbullet & -- & \textbullet & \textbullet & -- & -- & -- & -- \\
10 & \textsc{move}, scenes as drawn: base prompt\newline \textsc{move}, scenes as drawn: combined constraint\newline \textsc{move}, digits rearranged: base prompt\newline \textsc{move}, digits rearranged: combined constraint & \S6 & -- & -- & -- & -- & \textbullet & \textbullet & -- & -- & -- & -- \\
11 & rotated set: \textsc{mean}\newline rotated set: \textsc{move} & \S4 & \textbullet & \textbullet & \textbullet & \textbullet & \textbullet & \textbullet & \textbullet & \textbullet & -- & -- \\
12 & \textsc{mean}, \textsc{move}, ``want you to move'' and ``the block that you will move'', each with and without the visibility question\newline \textsc{move}: visible blocks stated\newline \textsc{move}: question after the request\newline \textsc{move}: question in a first turn & \S4 & -- & -- & \textbullet & \textbullet$^{b}$ & \textbullet & -- & -- & -- & -- & -- \\
13 & \textsc{mean} and \textsc{move}, each with and without the visibility question\newline \textsc{move}: block-count question\newline \textsc{move}: visible blocks stated\newline \textsc{move}: question after the request\newline \textsc{move}: question in a first turn & \S4 & -- & -- & -- & -- & -- & -- & -- & -- & \textbullet & -- \\
open & \textsc{mean}\newline \textsc{move}\newline \textsc{move}: without the visibility question\newline \textsc{move}: combined constraint & \S\S4--6 & -- & -- & -- & -- & -- & -- & -- & -- & \textbullet & \textbullet \\
\bottomrule
\end{tabular}
\caption{The batches (\S\ref{sec:design}) and the sections that compare their conditions.
\textbullet: the model answers them; $^{a}$:~grok-4.6 answers only \textsc{mean} and \textsc{move} in row~1, and 771 of the 800 calls of each; every other model answers every call.
$^{b}$:~in row~12 gpt-5.6-terra also answers the block-count question.
Row~10: the main set as drawn and a copy with the digits rearranged; row~11: the rotated set; other rows: the main set.
Rows~12 and~13: Table~\ref{tab:e13}.}
\label{tab:design}
\end{table*}

\paragraph{Collection.}
All calls to the closed models were made between 2026-09-17 and 2026-10-12 (UTC).
The closed models received each call at the provider's default temperature, without an output cap and at reasoning effort \texttt{medium}; no reply stopped at a length limit.
A reply counts as answered when, with any Markdown code fences removed, it contains a JSON object with the fields the prompt asks for, each of the type asked: text around the object is ignored, the last such object is read, a number in quotes is accepted, and a number not on the table is scored wrong.

\paragraph{Open-weight models.}
Qwen3.8-27B and Muse Glimmer-30B are, for their two families, the newest and largest dense open-weight vision-language models that the available hardware could serve.
Both are released as BF16 checkpoints, here at Hugging Face revisions \texttt{1d4bf0f} and \texttt{a4e59da}, which vLLM 0.31.0 served locally in float16, as the GPUs used do not support BF16, with each model's own chat template.
A request is one user message, the image followed by the prompt, with no system message; Muse Glimmer-30B's template adds a system block.
Both models reason before answering, at the medium level of each template's variable, and only the answer is parsed; sampling follows each model's generation settings.

\paragraph{Availability.}
The supplement holds the images and answer keys of both sets and of the copy with the digits rearranged, the position of every block and of the director's eye in the main set, every reply of the conditions in Table~\ref{tab:design} with its score, the codes and quotes of the 4,800 coded reasoning traces, the codes of the second pass in Appendix~\ref{app:traces}, and the two human coders' answer sheets.

\section{Stimuli and Answer Keys}
\label{app:stim}

Each scene places three or four cubes of different sizes, each with its own digit and colour, on a tabletop in front of an opaque partition, with a director figure behind the partition facing the model.
A block counts as hidden when every line of sight from it to the director's eye meets the partition, and as visible when none does.
A scene with a partly hidden block is discarded.
An independent verifier that shares no function with the generator renders the director's view with a depth buffer, and a scene is kept only when the two agree on every block.
Digits are drawn without replacement, independently of size.
In 40 of the 160 main-set scenes the hidden block carries the largest digit, and in 95 a block the director can see that is smaller than the director-view block carries a larger digit.
The answer keys give each cube's edge.
The rotated set holds 160 scenes of three blocks; the block hidden in a scene is the largest, the middle or the smallest, in 54, 53 and 53 scenes.

\paragraph{Apparent size from the director's side.}
The answer keys rank the blocks by edge, and from the director's eye a nearer block can look larger than a bigger block farther away.
Measured by the visual angle $2\arctan(e/2d)$ of a block's edge $e$, with $d$ the distance from the director's eye to the block's centre, a visible block other than the director-view block looks largest from the director's side in 19 of the 160 main-set scenes.
Figure~\ref{fig:example} shows the first scene in scene order that is not among them.
Recomputed without these scenes, the intervals in Table~\ref{tab:summary} for the changes with the visibility question and with the combined constraint exclude zero on the same models as on all scenes, with one addition: gpt-5.6-luna's change under \textsc{mean} becomes $-1.8$ $[-3.6, -0.1]$ points.
On the main set the interval for visibility accuracy under \textsc{mean} minus under \textsc{move} stays within the $\pm$10-point margin on every model, and on the scenes as drawn the joint rate under the combined constraint stays below 1\% in every batch.

\paragraph{Rearranged digits.}
A copy of the main set's 160 scenes rearranges each scene's digits, drawn at random among the arrangements that change both whether a visible block smaller than the director-view block carries a larger digit and whether the hidden block carries the largest digit.
Redrawn with the original digits the scenes match pixel for pixel; the changed pixels lie on the blocks, and both visibility checkers recompute the keys.

\section{Statistics}
\label{app:stats}

\paragraph{Estimators.}
Each comparison is fitted on the answered replies of its conditions, with the scene as the cluster.
Generalized estimating equations take $\operatorname{logit} P(y = 1) = \beta_0 + \beta_1 x$, with $x = 1$ for one condition of the pair and $0$ for the other, an exchangeable working correlation estimated by moments, and the cluster-robust sandwich covariance times $G/(G - 1)$ over $G$ clusters.
The difference of the fitted shares gets a delta-method $t$ interval on $G - 2$ degrees of freedom.
Where a condition has no event, the paired estimator takes the mean of the clusters' differences, with a $t$ interval.
Where every cluster gives the same difference $c$, a mean over clusters bounded by $l$ and $u$ per cluster gets $[c - (c - l)U, c + (u - c)U]$, and $[-U, +U]$ for differences at zero, with $U = 1 - 0.025^{1/G}$, 2.3 points on 160 scenes.
Each end fails with probability at most 0.025 whatever the correlation within a cluster.
The bound is of the rule-of-three kind: if a share $U$ of the clusters differed from $c$, all $G$ would show $c$ with probability 0.025, and each differing cluster moves the mean by at most the width of its range.
A single share has a percentile interval over 2000 scene resamples.
Rates and differences are rounded to four places, and figures in percent or points round them further.
At every precision an exact half is rounded away from zero.

\paragraph{The margin.}
The 10-point margin of \S\ref{sec:asking}, a change of one reply in ten, is small beside the changes that a first question makes to the answer to the second in comparable tasks: 30 to 33 points when SimpleToM gives each model its own answer to the awareness question \citep{gu2024simpletom}, and 47 points when the closest prior design asks first which blocks the director can see \citep{zhang2026crosstask}.
The accuracy differences tested against the margin are outside the families below, and tables do not mark them.

\paragraph{The corrections.}
Benjamini--Hochberg correction at a false discovery rate of 0.05 takes the comparisons of each section as one family, with two-sided $p$ from the same estimators: 72 in \S\ref{sec:asking}, 42 in \S\ref{sec:readings} and 39 in \S\ref{sec:constraint}.
Each family holds the comparisons of its section on every model asked them (Table~\ref{tab:design}).
The family of \S\ref{sec:asking} holds the change with the visibility question under \textsc{move}, the same change in each of the two digit strata, the block-count question, \textsc{mean} with and without the visibility question, and the comparisons of Table~\ref{tab:e13}.
The family of \S\ref{sec:readings} holds, for each open-weight model, the changes with the visibility question in the three readings, in the hidden-block choices whose trace takes the director not to see the block and in the traces that work out the director's view, and \textsc{move} minus \textsc{mean} in the three readings; each rewording minus \textsc{move}; and the knowledge question minus the visibility question, among all replies and among those whose report leaves the hidden block out.
The family of \S\ref{sec:constraint} holds the combined constraint minus the base prompt in the joint rate and in the hidden-block choice rate, the change with the visibility question in the hidden-block choice rate under each of the two prompts, the padding sentence and each clause alone in the joint rate, and gpt-6-luna's visibility accuracy in rows~6, 8 and~9 of Table~\ref{tab:design}.
Two estimates in Table~\ref{tab:e13} have an interval that excludes zero and an adjusted $p$ of 0.058, above 0.05, and are marked $^{\dagger}$: gpt-5.6-luna's change under \textsc{mean} and gpt-6-sol's change under ``want you to move'', each with minus without the visibility question.

\section{Estimates and Counts}
\label{app:counts}

Figure~\ref{fig:logic}'s right column names the table that holds each row's estimates.
Table~\ref{tab:knowledge} gives the knowledge question of \S\ref{sec:readings}, and Table~\ref{tab:parts} the parts and controls of the combined constraint of \S\ref{sec:constraint}.
The other tables give the counts behind Figures~\ref{fig:asking}--\ref{fig:constraint} and \S\S\ref{sec:asking}--\ref{sec:constraint}.

\begin{table*}[t]
\centering
\scriptsize
\setlength{\tabcolsep}{2.6pt}
\begin{tabular}{@{}lcccccccccccc@{}}
\toprule
 & \multicolumn{4}{c}{\textsc{move}, without the visibility question} & \multicolumn{4}{c}{\textsc{move}, with the
visibility question} & \multicolumn{4}{c}{\textsc{move}, combined constraint} \\
\cmidrule(lr){2-5}\cmidrule(lr){6-9}\cmidrule(l){10-13}
model & \dci{director-}{view} & hidden & \dci{largest-digit}{visible} & other & \dci{director-}{view} & hidden &
\dci{largest-digit}{visible} & other & \dci{director-}{view} & hidden & \dci{largest-digit}{visible} & other \\
\midrule
gpt-5.5 & 92.9 & 2.0 & 4.4 & 0.8 & 79.9 & 13.8 & 5.6 & 0.8 & 86.3 & 1.4 & 12.0 & 0.4 \\
gpt-5.6-sol & 90.6 & 2.0 & 7.3 & 0.1 & 81.0 & 13.8 & 4.9 & 0.4 & 84.8 & 1.4 & 13.8 & 0.1 \\
gpt-6-sol & 91.1 & 6.5 & 2.1 & 0.3 & 46.8 & 52.5 & 0.3 & 0.5 & 96.1 & 0.9 & 3.0 & 0.0 \\
gpt-5.6-terra & 90.8 & 3.6 & 4.8 & 0.9 & 6.4 & 91.9 & 1.3 & 0.5 & 70.0 & 1.5 & 27.4 & 1.1 \\
gpt-5.6-luna & 60.3 & 4.8 & 32.1 & 2.9 & 7.6 & 82.1 & 10.1 & 0.1 & 52.6 & 1.5 & 44.3 & 1.6 \\
gpt-6-luna & 36.0 & 37.3 & 24.6 & 2.1 & 7.9 & 78.3 & 13.9 & 0.0 & 55.6 & 7.9 & 35.1 & 1.4 \\
grok-4.6 & 86.3 & 3.0 & 8.4 & 2.4 & 87.9 & 2.4 & 6.3 & 3.5 & -- & -- & -- & -- \\
Qwen3.8-27B & 64.6 & 24.9 & 9.4 & 1.1 & 51.4 & 41.8 & 6.4 & 0.5 & 86.8 & 2.5 & 10.6 & 0.1 \\
Muse Glimmer-30B & 56.0 & 20.8 & 20.0 & 3.3 & 55.8 & 20.4 & 20.8 & 3.1 & 43.9 & 18.1 & 36.1 & 1.9 \\
\bottomrule
\end{tabular}
\caption{Block choices as percentages of the answered \textsc{move} replies: the director-view block, the hidden block, the largest-digit visible block and any other block (other).
Without and with the visibility question come from one batch per model, the combined constraint for GPT models from another; --:~not asked.}
\label{tab:choices}
\end{table*}

\begin{table*}[t]
\centering
\scriptsize
\setlength{\tabcolsep}{3pt}
\begin{tabular}{@{}lccccc@{}}
\toprule
 & \multicolumn{2}{c}{\dci{hidden-block choice rate:}{with minus without the visibility question}} &
\multicolumn{2}{c}{\dci{visibility accuracy:}{\textsc{mean} minus \textsc{move}}} &
\tci{joint rate, \textsc{move}:}{combined constraint}{minus base prompt} \\
\cmidrule(lr){2-3}\cmidrule(lr){4-5}
model & \textsc{move} & \textsc{mean} & main set & rotated set & \\
\midrule
gpt-5.5 & $+11.8$ $[+8.8, +14.8]^{*}$ & $-1.5$ $[-2.8, -0.2]^{*}$ & $-0.4$ $[-2.3, +1.5]$ & $+0.8$ $[-1.7, +3.2]$ & $-11.3$ $[-13.7, -8.8]^{*}$ \\
gpt-5.6-sol & $+11.8$ $[+9.1, +14.4]^{*}$ & -- & $+0.8$ $[-1.2, +2.7]$ & $-0.4$ $[-2.6, +1.9]$ & $-15.1$ $[-17.8, -12.5]^{*}$ \\
gpt-6-sol & $+46.0$ $[+41.4, +50.6]^{*}$ & $-0.3$ $[-1.0, +0.5]$ & $+1.3$ $[-0.1, +2.6]$ & $+1.0$ $[-0.5, +2.5]$ & $-46.0$ $[-50.1, -41.9]^{*}$ \\
gpt-5.6-terra & $+88.3$ $[+85.3, +91.2]^{*}$ & -- & $+0.1$ $[-2.1, +2.4]$ & $+0.3$ $[-2.4, +2.9]$ & $-82.9$ $[-86.3, -79.5]^{*}$ \\
gpt-5.6-luna & $+77.4$ $[+73.3, +81.4]^{*}$ & $-1.4$ $[-3.0, +0.2]$ & $+1.3$ $[-1.2, +3.7]$ & $+3.1$ $[+0.5, +5.8]$ & $-72.5$ $[-76.9, -68.1]^{*}$ \\
gpt-6-luna & $+41.0$ $[+35.2, +46.9]^{*}$ & $-9.6$ $[-14.0, -5.2]^{*}$ & $+1.8$ $[-0.9, +4.4]$ & $-0.6$ $[-3.4, +2.2]$ & $-72.8$ $[-77.4, -68.1]^{*}$ \\
gpt-6-astra & -- & -- & $+0.5$ $[-0.3, +1.3]$ & $-0.3$ $[-1.0, +0.5]$ & -- \\
grok-4.6 & $-0.6$ $[-2.1, +0.8]$ & -- & $+1.1$ $[-1.3, +3.5]$ & $-0.3$ $[-3.1, +2.6]$ & -- \\
Qwen3.8-27B & $+16.9$ $[+11.8, +22.0]^{*}$ & -- & $-1.6$ $[-3.6, +0.4]$ & -- & $-38.4$ $[-42.3, -34.5]^{*}$ \\
Muse Glimmer-30B & $-0.4$ $[-3.1, +2.3]$ & -- & $-0.5$ $[-4.4, +3.4]$ & -- & $-0.3$ $[-0.6, +0.1]$ \\
\bottomrule
\end{tabular}
\caption{Appendix~\ref{app:counts}: main estimates in points with 95\% intervals, each within one batch; visibility accuracy under the base prompt.
$^{*}$:~an increase or a decrease by the rule of \S\ref{sec:design}; accuracy differences are tested against the $\pm$10-point margin and not marked (Appendix~\ref{app:stats}); --:~not asked.}
\label{tab:summary}
\end{table*}

\begin{table*}[t]
\centering
\scriptsize
\setlength{\tabcolsep}{3pt}
\begin{tabular}{@{}lcccccc@{}}
\toprule
 & \multicolumn{4}{c}{\textsc{move}} & \multicolumn{2}{c}{\textsc{mean}} \\
\cmidrule(lr){2-5}\cmidrule(l){6-7}
model & \dci{visibility}{question} & \dci{block-count}{question} & \dci{without the}{visibility question} &
\dci{block-count question}{minus without} & \dci{visibility}{question} & \dci{without the}{visibility question} \\
\midrule
gpt-5.5 & 110 & 19 & 16 & $+0.4$ $[-0.9, +1.6]$ & 9 & 21 \\
gpt-5.6-sol & 110 & -- & 16 & -- & -- & -- \\
gpt-6-sol & 420 & 38 & 52 & $-1.8$ $[-4.1, +0.6]$ & 6 & 8 \\
gpt-5.6-terra & 735 & -- & 29 & -- & -- & -- \\
gpt-5.6-luna & 657 & 56 & 38 & $+2.3$ $[-0.02, +4.5]$ & 13 & 24 \\
gpt-6-luna & 626 & 341 & 298 & $+5.4$ $[+0.5, +10.3]^{*}$ & 93 & 170 \\
grok-4.6 & 19 & 26 & 24 & $+0.3$ $[-1.2, +1.7]$ & -- & -- \\
Qwen3.8-27B & 334 & -- & 199 & -- & -- & -- \\
Muse Glimmer-30B & 163 & -- & 166 & -- & -- & -- \\
\bottomrule
\end{tabular}
\caption{\S\ref{sec:asking}: hidden-block choices, whatever the visibility report, of 800 answered replies, and the difference in points with its 95\% interval.
\textsc{move}: rows~2 and~3 of Table~\ref{tab:design} and the open-weight models' batch; \textsc{mean}: row~4.
Interval ends that round to zero at one decimal are given to two.}
\label{tab:askcounts}
\end{table*}

\begin{table*}[t]
\centering
\scriptsize
\setlength{\tabcolsep}{2.4pt}
\begin{tabular}{@{}lcccccc@{}}
\toprule
 & \multicolumn{3}{c}{\dci{hidden block with}{the largest digit}} & \multicolumn{3}{c}{\dci{hidden block without}{the largest digit}} \\
\cmidrule(lr){2-4}\cmidrule(l){5-7}
model & with & without & difference & with & without & difference \\
\midrule
gpt-5.5 & 27 & 4 & $+11.5$ $[+5.5, +17.5]^{*}$ & 83 & 12 & $+11.8$ $[+8.3, +15.4]^{*}$ \\
gpt-5.6-sol & 20 & 2 & $+9.0$ $[+4.4, +13.6]^{*}$ & 90 & 14 & $+12.7$ $[+9.5, +15.8]^{*}$ \\
gpt-6-sol & 101 & 14 & $+43.5$ $[+34.9, +52.1]^{*}$ & 319 & 38 & $+46.8$ $[+41.3, +52.3]^{*}$ \\
gpt-5.6-terra & 190 & 4 & $+93.0$ $[+88.3, +97.7]^{*}$ & 545 & 25 & $+86.7$ $[+83.0, +90.3]^{*}$ \\
gpt-5.6-luna & 199 & 11 & $+94.0$ $[+89.6, +98.4]^{*}$ & 458 & 27 & $+71.8$ $[+67.0, +76.7]^{*}$ \\
gpt-6-luna & 200 & 152 & $+24.0$ $[+15.4, +32.6]^{*}$ & 426 & 146 & $+46.7$ $[+39.7, +53.7]^{*}$ \\
grok-4.6 & 4 & 6 & $-1.0$ $[-4.2, +2.2]$ & 15 & 18 & $-0.5$ $[-2.2, +1.2]$ \\
Qwen3.8-27B & 91 & 52 & $+19.5$ $[+9.6, +29.4]^{*}$ & 243 & 147 & $+16.0$ $[+10.0, +22.0]^{*}$ \\
Muse Glimmer-30B & 65 & 57 & $+4.0$ $[-1.8, +9.8]$ & 98 & 109 & $-1.8$ $[-4.9, +1.2]$ \\
\bottomrule
\end{tabular}
\caption{\S\ref{sec:asking}: \textsc{move} hidden-block choices with and without the visibility question on the 40 main-set scenes in which the hidden block carries the largest digit (of 200 replies) and on the other 120 (of 600), with the difference in points and its 95\% interval; the batches of Table~\ref{tab:askcounts}.}
\label{tab:strata}
\end{table*}

\begin{table*}[t]
\centering
\scriptsize
\setlength{\tabcolsep}{2.4pt}
\begin{tabular}{@{}lcccccc@{}}
\toprule
model & \textsc{mean} & \dci{``want you}{to move''} & \textsc{move} & \dci{``the block that}{you will move''} & \dci{``want'' minus}{\textsc{move}} & \dci{``the block'' minus}{\textsc{move}} \\
\midrule
gpt-5.5 & 11 & 10 & 109 & 359 & $-12.4$ $[-15.3, -9.4]^{*}$ & $+31.3$ $[+27.3, +35.2]^{*}$ \\
gpt-5.6-sol & 8 & 15 & 126 & 340 & $-13.9$ $[-16.9, -10.9]^{*}$ & $+26.8$ $[+22.1, +31.4]^{*}$ \\
gpt-6-sol & 4 & 6 & 412 & 545 & $-50.8$ $[-54.9, -46.6]^{*}$ & $+16.6$ $[+11.9, +21.4]^{*}$ \\
gpt-5.6-terra & 63 & 122 & 730 & 732 & $-76.0$ $[-79.4, -72.6]^{*}$ & $+0.3$ $[-2.2, +2.7]$ \\
gpt-5.6-luna & 12 & 354 & 681 & 715 & $-40.9$ $[-45.3, -36.5]^{*}$ & $+4.3$ $[+1.1, +7.4]^{*}$ \\
gpt-6-luna & 85 & 505 & 614 & 646 & $-13.6$ $[-18.1, -9.1]^{*}$ & $+4.0$ $[-0.2, +8.2]$ \\
gpt-6-astra & 1 & 1 & 2 & 4 & $-0.1$ $[-0.4, +0.1]$ & $+0.3$ $[-0.1, +0.6]$ \\
\bottomrule
\end{tabular}
\caption{\S\ref{sec:readings}: hidden-block choices, whatever the visibility report, of 800 answered replies with the base prompt, under \textsc{mean}, \textsc{move} and the two rewordings of \textsc{move}, row~1 of Table~\ref{tab:design}, and the differences in points with 95\% intervals.}
\label{tab:rewordings}
\end{table*}

\begin{table*}[t]
\centering
\scriptsize
\setlength{\tabcolsep}{3pt}
\begin{tabular}{@{}lccccccc@{}}
\toprule
 & \multicolumn{2}{c}{joint rate} & \multicolumn{2}{c}{hidden-block choices} & \multicolumn{2}{c}{correct reports} & \dci{combined: hidden block}{listed as visible} \\
\cmidrule(lr){2-3}\cmidrule(lr){4-5}\cmidrule(lr){6-7}
model & base & combined & base & combined & base & combined & \\
\midrule
gpt-5.5 & 90 & 0 & 105 & 11 & 771 & 769 & 11 \\
gpt-5.6-sol & 121 & 0 & 141 & 11 & 754 & 764 & 11 \\
gpt-6-sol & 368 & 0 & 384 & 7 & 775 & 782 & 7 \\
gpt-5.6-terra & 663 & 0 & 725 & 12 & 733 & 738 & 12 \\
gpt-5.6-luna & 581 & 1 & 659 & 12 & 706 & 729 & 11 \\
gpt-6-luna & 582 & 0 & 621 & 63 & 741 & 668 & 63 \\
Qwen3.8-27B & 307 & 0 & 334 & 20 & 764 & 756 & 20 \\
Muse Glimmer-30B & 2 & 0 & 163 & 145 & 518 & 491 & 145 \\
\bottomrule
\end{tabular}
\caption{Figure~\ref{fig:constraint} and \S\ref{sec:constraint}: \textsc{move} under the base prompt (base) and the combined constraint (combined), row~6 of Table~\ref{tab:design} and the open-weight models' batch, counts of 800 answered replies.}
\label{tab:constraintcounts}
\end{table*}

\begin{table*}[t]
\centering
\scriptsize
\setlength{\tabcolsep}{2pt}
\begin{tabular}[t]{@{}lccc@{}}
\toprule
\multicolumn{4}{@{}l}{(a) each clause alone: joint rate} \\
\midrule
model & base & \dci{knowledge}{constraint} & \dci{reference}{constraint} \\
\midrule
gpt-5.5 & 13.1 & 0.0 & 0.0 \\
gpt-5.6-sol & 12.4 & 0.0 & 0.0 \\
gpt-6-sol & 50.0 & 0.8 & 0.0 \\
gpt-5.6-terra & 81.6 & 0.4 & 3.5 \\
gpt-5.6-luna & 75.5 & 16.3 & 3.4 \\
gpt-6-luna & 73.8 & 6.1 & 10.9 \\
\bottomrule
\end{tabular}
\hfill
\begin{tabular}[t]{@{}lcccc@{}}
\toprule
\multicolumn{5}{@{}l}{(b) the padding sentence: joint rate} \\
\midrule
model & base & \dci{padding}{sentence} & combined & \dci{padding}{minus base} \\
\midrule
gpt-5.6-luna & 75.1 & 74.1 & 0.0 & $-1.0$ $[-5.1, +3.1]$ \\
gpt-6-luna & 73.0 & 59.8 & 0.1 & $-13.3$ $[-18.1, -8.5]^{*}$ \\
\bottomrule
\end{tabular}
\hfill
\begin{tabular}[t]{@{}lccc@{}}
\toprule
\multicolumn{4}{@{}l}{(c) digits rearranged: joint rate} \\
\midrule
model & \dci{drawn,}{combined} & \dci{rearranged,}{base} & \dci{rearranged,}{combined} \\
\midrule
gpt-5.6-luna & 0.0 & 84.6 & 0.0 \\
gpt-6-luna & 0.3 & 84.0 & 0.8 \\
\bottomrule
\end{tabular}

\vspace{6pt}
\renewcommand{\dci}[2]{#1~#2}
\begin{tabular}{@{}lccccccc@{}}
\toprule
\multicolumn{8}{@{}l}{(d) the visibility question under each prompt} \\
\midrule
 & \multicolumn{4}{c}{hidden-block choices} & joint & \multicolumn{2}{c}{hidden-block choices, with minus without} \\
\cmidrule(lr){2-5}\cmidrule(lr){6-6}\cmidrule(l){7-8}
model & base, with & base, without & combined, with & combined, without & combined, with & base & combined \\
\midrule
gpt-6-sol & 56.1 & 8.6 & 0.9 & 1.1 & 0.0 & $+47.5$ $[+42.8, +52.2]^{*}$ & $-0.3$ $[-0.9, +0.4]$ \\
gpt-5.6-luna & 85.0 & 4.0 & 1.3 & 2.1 & 0.0 & $+81.0$ $[+77.3, +84.7]^{*}$ & $-0.9$ $[-1.8, +0.01]$ \\
gpt-6-luna & 77.9 & 39.6 & 9.5 & 6.5 & 0.1 & $+38.3$ $[+32.6, +43.9]^{*}$ & $+3.0$ $[-0.2, +6.2]$ \\
\bottomrule
\end{tabular}
\caption{\S\ref{sec:constraint}: parts and controls of the combined constraint, \textsc{move}, each panel from one batch per model: percentages of the answered replies and differences in points with 95\% intervals; base: base prompt; combined: combined constraint; with, without: the visibility question; joint: joint rate.
Ends that round to zero are given to two places.}
\label{tab:parts}
\end{table*}

\begin{table*}[t]
\centering
\scriptsize
\setlength{\tabcolsep}{1.6pt}
\begin{tabular}{@{}lccccccccc@{}}
\toprule
 & \multicolumn{5}{c}{reading} & \dci{works out the}{director's view} & \multicolumn{3}{c}{\tci{hidden-block choices:}{chosen block,}{per the trace}} \\
\cmidrule(lr){2-6}\cmidrule(l){8-10}
condition & \dci{director}{view} & \dci{own}{view} & \dci{prior}{knowledge} & \dci{largest}{digit} & \dci{no}{rule} & & seen & unseen & \dci{not}{stated} \\
\midrule
\multicolumn{10}{@{}l}{\textit{Qwen3.8-27B}} \\
\textsc{move}, with & 402 & 312 & 16 & 70 & 0 & 800 & 16 & 318 & 0 \\
\textsc{move}, without & 593 & 99 & 1 & 107 & 0 & 751 & 114 & 2 & 83 \\
\textsc{mean} & 728 & 1 & 1 & 70 & 0 & 800 & 26 & 2 & 0 \\
with minus without & \dci{$-23.9$}{$[-28.7, -19.0]^{*}$} & \dci{$+26.6$}{$[+22.1, +31.1]^{*}$} & \dci{$+1.9$}{$[+0.8, +2.9]$} & \dci{$-4.6$}{$[-7.8, -1.5]^{*}$} & \dci{$0.0$}{$[-2.3, +2.3]$} & \dci{$+6.1$}{$[+4.5, +7.8]^{*}$} &  & \dci{$+39.5$}{$[+35.6, +43.4]^{*}$} &  \\
with minus \textsc{mean} & \dci{$-40.8$}{$[-45.1, -36.4]^{*}$} & \dci{$+38.9$}{$[+34.9, +42.9]^{*}$} & \dci{$+1.9$}{$[+0.8, +2.9]$} & \dci{$0.0$}{$[-2.5, +2.5]$} & \dci{$0.0$}{$[-2.3, +2.3]$} & \dci{$0.0$}{$[-2.3, +2.3]$} &  &  &  \\
\multicolumn{10}{@{}l}{\textit{Muse Glimmer-30B}} \\
\textsc{move}, with & 599 & 9 & 0 & 192 & 0 & 800 & 160 & 3 & 0 \\
\textsc{move}, without & 618 & 1 & 0 & 180 & 1 & 800 & 163 & 0 & 3 \\
\textsc{mean} & 689 & 2 & 0 & 109 & 0 & 800 & 177 & 0 & 0 \\
with minus without & \dci{$-2.4$}{$[-5.5, +0.8]$} & \dci{$+1.0$}{$[+0.2, +1.8]^{*}$} & \dci{$0.0$}{$[-2.3, +2.3]$} & \dci{$+1.5$}{$[-1.5, +4.5]$} & \dci{$-0.1$}{$[-0.4, +0.1]$} & \dci{$0.0$}{$[-2.3, +2.3]$} &  & \dci{$+0.4$}{$[-0.1, +0.8]$} &  \\
with minus \textsc{mean} & \dci{$-11.3$}{$[-14.5, -8.0]^{*}$} & \dci{$+0.9$}{$[+0.1, +1.7]^{*}$} & \dci{$0.0$}{$[-2.3, +2.3]$} & \dci{$+10.4$}{$[+7.2, +13.5]^{*}$} & \dci{$0.0$}{$[-2.3, +2.3]$} & \dci{$0.0$}{$[-2.3, +2.3]$} &  &  &  \\
\bottomrule
\end{tabular}
\caption{Appendix~\ref{app:traces}: the codes of the 4,800 reasoning traces, counts of 800 per condition, and for each model the differences between conditions in points with 95\% intervals; differences outside the corrected families of Appendix~\ref{app:stats}, those of the prior-knowledge and no-rule readings and the works-out difference with minus \textsc{mean}, are not marked.}
\label{tab:traces}
\end{table*}

\begin{table}[t]
\centering
\scriptsize
\setlength{\tabcolsep}{1.6pt}
\begin{tabular}{@{}lcccccc@{}}
\toprule
 & \multicolumn{2}{c}{\dci{visibility}{question}} & \multicolumn{2}{c}{\dci{knowledge}{question}} &
\multicolumn{2}{c}{\dci{knowledge minus}{visibility}} \\
\cmidrule(lr){2-3}\cmidrule(lr){4-5}\cmidrule(l){6-7}
model & all & \dci{left}{out} & all & \dci{left}{out} & all & \dci{left}{out} \\
\midrule
gpt-5.5 & 13.9 & 12.8 & 5.5 & 4.3 & \dci{$-8.4$}{$[-11.4, -5.3]^{*}$} & \dci{$-8.5$}{$[-11.4, -5.6]^{*}$} \\
gpt-5.6-sol & 13.6 & 12.6 & 6.5 & 3.9 & \dci{$-7.1$}{$[-10.2, -4.1]^{*}$} & \dci{$-8.8$}{$[-11.4, -6.1]^{*}$} \\
gpt-6-sol & 54.3 & 53.8 & 36.5 & 36.0 & \dci{$-17.8$}{$[-22.5, -13.0]^{*}$} & \dci{$-17.8$}{$[-22.5, -13.0]^{*}$} \\
gpt-5.6-terra & 92.1 & 92.1 & 6.1 & 4.6 & \dci{$-86.0$}{$[-89.0, -83.1]^{*}$} & \dci{$-87.6$}{$[-90.3, -84.8]^{*}$} \\
gpt-5.6-luna & 84.6 & 84.9 & 39.4 & 38.1 & \dci{$-45.3$}{$[-50.0, -40.5]^{*}$} & \dci{$-46.8$}{$[-51.7, -42.0]^{*}$} \\
gpt-6-luna & 76.4 & 76.9 & 47.9 & 47.4 & \dci{$-28.5$}{$[-33.1, -23.9]^{*}$} & \dci{$-29.5$}{$[-34.4, -24.6]^{*}$} \\
\bottomrule
\end{tabular}
\caption{\S\ref{sec:readings}: \textsc{move} hidden-block choices with the visibility question and with the knowledge question in its place, in one batch per model: percentages of the answered replies (all) and of the 759 to 794 replies whose report leaves the hidden block out (left out), and the differences between the two questions in points with 95\% intervals.}
\label{tab:knowledge}
\end{table}

\begin{table}[t]
\centering
\scriptsize
\setlength{\tabcolsep}{2.4pt}
\begin{tabular}{@{}lrrrrrrrrrc@{}}
\toprule
 & \multicolumn{9}{c}{row of Table~\ref{tab:design}} & \\
\cmidrule(lr){2-10}
model & 1 & 2 & 3 & 5 & 6 & 7 & 8 & 9 & 10 & range \\
\midrule
gpt-5.5 & 109 & 110 & -- & 111 & 105 & 119 & -- & -- & -- & 105--119 \\
gpt-5.6-sol & 126 & -- & 110 & 109 & 141 & 112 & -- & -- & -- & 109--141 \\
gpt-6-sol & 412 & 420 & -- & 434 & 384 & 414 & -- & 449 & -- & 384--449 \\
gpt-5.6-terra & 730 & -- & 735 & 737 & 725 & 718 & -- & -- & -- & 718--737 \\
gpt-5.6-luna & 681 & 657 & -- & 677 & 659 & 684 & 666 & 680 & 673 & 657--684 \\
gpt-6-luna & 614 & 626 & -- & 611 & 621 & 630 & 622 & 623 & 623 & 611--630 \\
\bottomrule
\end{tabular} {}
\caption{\S\ref{sec:design}: \textsc{move} hidden-block choices under the base prompt, of 800 answered replies, in each batch that asks it; gpt-6-astra answers it in row~1 only.}
\label{tab:batches}
\end{table}

\section{Coding the Reasoning Traces}
\label{app:traces}

\paragraph{Codebook.}
The codebook asks what the reasoning says and believes, not what is true in the picture, and the coder sees neither the picture nor any answer key.
The first code is the reading behind the final block choice, one of five.
A trace has the director-view reading when it picks the largest block among those it takes the director to see, or checks its pick against the director's view.
It has the own-view reading when it picks the largest block among all the blocks as the model sees them, without limiting the choice to the director's view or giving the director's knowledge as a reason.
It has the prior-knowledge reading when it picks a block it takes the director not to see now and gives as the reason that the director knows of that block.
It has the largest-digit reading when it takes ``largest'' to mean the largest digit, among all the blocks or those the director can see.
Otherwise no rule is identifiable.
Where size and digit are both given as reasons for the same block, the code goes by size; the last rule applied before the reply counts.
The second code records whether, by its last stated belief, the trace takes the director to see the chosen block: seen, unseen or not stated.
Where the trace states no belief and the reply has a visibility report, the code follows that report.
A third code records whether the trace, anywhere, works out which blocks the director can or cannot see.
Each code comes with a quote of at most 40 words copied word for word from the trace.

\paragraph{Coder and checks.}
The coder, gpt-6-astra at maximum reasoning effort, received batches of traces with the codebook and only each item's neutral identifier, reply and reasoning, not the model, the condition or the scene.
On 30 test traces written to a known code, mixed with 30 real items, it gave the intended reading and second code on all 30.
The real items, coded a second time in the main run, received the same two codes on all 30.
It coded all 4,800 traces of \textsc{move} with and without the visibility question and of \textsc{mean} on the main set.

\paragraph{The second code without the report.}
To test whether the second code of the hidden-block choices rests on the report, the coder coded Qwen3.8-27B's hidden-block choices under \textsc{move} a second time, shown only a new neutral identifier, the chosen block and the trace, with a trace that states no belief about that block coded as not saying.
All 334 choices with the visibility question keep their code, among them the 318 whose trace takes the director not to see the block.
Of the 199 choices without the question, whose replies carry no report, 195 keep theirs (Cohen's $\kappa$ 0.96).
In this pass the coder gave the intended code on all 32 test traces written to a known code.
The codes of the hidden-block choices with the question thus do not depend on the report.

\paragraph{Human check.}
Two people coded, each on their own, the same 42 items by the same codebook.
Of these, 36 traces were drawn three from each of 12 strata (the two models, \textsc{move} with and without the visibility question, and the director-view block, the hidden block or any other block as the choice), without regard to any automatic code and with no model and scene twice.
The other 6, mixed in among them, were test traces written to a known code, not marked as such.
The two coders did not know the hypotheses of the study, and they saw neither the model, the condition, the scene nor the automatic codes.
Both gave the intended codes on all 6 test traces.
On the 36 traces, their reading codes agree with the automatic codes on 36 and 34 (Cohen's $\kappa$ 1.00 $[1.00, 1.00]$ and 0.90 $[0.74, 1.00]$), and with each other on 34 ($\kappa$ 0.90 $[0.74, 1.00]$; 95\% intervals over 2000 resamples of the 36 traces).
Whether the trace takes the director to see the chosen block agrees on 36 in every pair.

\end{document}